\documentclass[10 pt, conference]{ieeeconf}

\IEEEoverridecommandlockouts
                           
\usepackage{graphicx}
\usepackage{amsmath,amssymb,amsfonts}
\usepackage{subcaption}
\usepackage{color}
\usepackage{placeins}

\usepackage{breakurl}
\usepackage{fixltx2e}
\usepackage{subcaption} 
\usepackage{lipsum}
\usepackage[ruled,vlined]{algorithm2e}
\usepackage{scalerel}
\usepackage{tikz}

\edef\normalE{\the\mathcode`E}
\begingroup\lccode`~=`E \lowercase{\endgroup
  \def~}{\mathbb{\mathchar\normalE}}
\mathcode`E="8000
\usepackage{hyperref} 

\begin{document}
\color{red}
\onecolumn
\huge 
\begin{center}
%2nd Conference on Blockchain Research & Applications for Innovative Networks and Services
\end{center}
\vskip 1.5in
\Large 

\vskip 1.5in
\Large 
"\textcopyright 2020 IEEE.  Personal use of this material is permitted.  Permission from IEEE must be obtained for all other uses, in any current or future media, including reprinting/republishing this material for advertising or promotional purposes, creating new collective works, for resale or redistribution to servers or lists, or reuse of any copyrighted component of this work in other works."

\color{black}
\twocolumn

\title{\LARGE \bf
Generative Adversarial Networks for Bitcoin Data Augmentation
}

\author{
\authorblockN{Francesco Zola\authorrefmark{1}, Jan Lukas Bruse\authorrefmark{1}, Xabier Etxeberria Barrio\authorrefmark{1}, Mikel Galar\authorrefmark{2}, Raul Orduna Urrutia\authorrefmark{1}}
\authorblockA{\authorrefmark{1} Vicomtech Foundation, Basque Research and Technology Alliance (BRTA)\\ Paseo Mikeletegi 57, 20009 Donostia/San Sebastian, Spain\\ \{fzola, jbruse, xetxeberria, rorduna\}@vicomtech.org \\
\authorrefmark{2}Institute of Smart Cities, Public University of Navarre, 31006 Pamplona, Spain \\ mikel.galar@unavarra.es }
}

\maketitle

\thispagestyle{empty}
\pagestyle{empty}

%%%%%%%%%%%%%%%%%%%%%%%%%%%%%%%%%%%%%%%%%%%%%%%%%%%%%%%%%%%%%%%%%%%%%%%%%%%%%%%%
\begin{abstract}

In Bitcoin entity classification, results are strongly conditioned by the ground-truth dataset, especially when applying supervised machine learning approaches.
However, these ground-truth datasets are frequently affected by significant class imbalance as generally they contain much more information regarding legal services (Exchange, Gambling), than regarding services that may be related to illicit activities (Mixer, Service). Class imbalance increases the complexity of applying machine learning techniques and reduces the quality of classification results, especially for underrepresented, but critical classes.

In this paper, we propose to address this problem by using Generative Adversarial Networks (GANs) for Bitcoin data augmentation as GANs recently have shown promising results in the domain of image classification. However, there is no ``one-fits-all" GAN solution that works for every scenario. In fact, setting GAN training parameters is non-trivial and heavily affects the quality of the generated synthetic data. We therefore evaluate how GAN parameters such as the optimization function, the size of the dataset and the chosen batch size affect GAN implementation for one underrepresented entity class (Mining Pool) and demonstrate how a ``good" GAN configuration can be obtained that achieves high similarity between synthetically generated and real Bitcoin address data. 
To the best of our knowledge, this is the first study presenting GANs as a valid tool for generating synthetic address data for data augmentation in Bitcoin entity classification.

\end{abstract}

\IEEEoverridecommandlockouts

%%%%%%%%%%%%%%%%%%%%%%%%%%%%%%%%%%%%%%%%%%%%%%%%%%%%%%%%%%%%%%%%%%%%%%%%%%%%%%%%
\section{Introduction}

Bitcoin (or BTC) is a cryptocurrency based on a publicly shared ledger called blockchain \cite{nakamoto2019bitcoin}. All transactions are stored in blocks of the blockchain that cannot be manipulated or changed \cite{crosby2016blockchain}. The access to this information is free for each user belonging to the Bitcoin network, while Bitcoin user identity is protected by anonymity.
Decreasing anonymity in the Bitcoin network has become a challenge when trying to discover new entities in the network \cite{harlev2018breaking}, or when aiming to detect entities related to illicit or abnormal activities in order to improve the trustworthiness within the network.

For Bitcoin entity classification, results are strongly conditioned by the ground-truth dataset, especially when using supervised machine learning approaches. Usually, public datasets or data generated by scraping forums and web-sites are used. Nevertheless, due to the complexity of the Bitcoin network and its anonymity policy, these datasets are typically characterized by heavily imbalanced classes of Bitcoin entities - with some classes being highly underrepresented compared to others. Such class imbalance affects the quality of a learning system because, once trained using imbalanced classes, learning algorithms are conditioned to resolve complicated classification problems based on a skewed class distribution and thus fail to detect underrepresented classes well \cite{fernandez2018learning}.

The class imbalance problem becomes even more relevant for restricted datasets for which it is difficult to detect and add new information, such as the Bitcoin blockchain. In fact, there is generally much more information available pertaining to legal services that do not need to obscure their information (such as Exchanges or legal Markets) than for entities that intend to mask the traces of their services (such as Mixers or Ransomware), which are often related to illicit activities. This calls for novel approaches that can augment the dataset of such critical services with synthetically created instances, ultimately aiming to improve classification results. The most common technique currently adopted to address the imbalance problem is to not consider certain classes \cite{zola2019bitcoin}, or to apply various types of sampling methods, like over-sampling for the least represented classes \cite{harlev2018breaking} or under-sampling for the most represented ones \cite{liu2008exploratory}.

To the best of our knowledge, this is the first work that addresses Bitcoin class imbalance by introducing address data augmentation using generative adversarial networks (GANs). The key idea behind GANs is to create synthetic data that cannot be distinguished from real data. The ``adversarial" aspect is introduced by using two algorithms/networks working against each other in order to improve their ability to learn and reproduce a real input dataset. The potential to learn and copy almost every dataset distribution has led researchers to apply GANs predominantly in the domain of image processing. Further, GANs can work with multi-modal outputs and can be trained to predict missing data \cite{goodfellow2016nips}.

Here, these concepts are leveraged in order to take a step forward towards resolving the imbalance problem related to Bitcoin entity classification. Setting GAN training parameters is non-trivial and heavily affects the quality of the generated synthetic data \cite{goodfellow2016nips}. Therefore, the aim of this paper is to study how GAN configuration parameters affect the produced synthetic data and which configuration should be used for achieving "best" data augmentation i.e. for generating synthetic data as similar as possible to real data. To investigate the impact of GAN configurations, several tests changing three important GAN parameters were carried out: the optimizer function, the size of the real input dataset and the batch size. In this manner, we were able to evaluate how each variable affects the training and the generation phase. Moreover, we present how a ``good" GAN configuration can be obtained that allows for efficient generation of synthetic address data.

The initial Bitcoin blockchain dataset used in this work was composed by entities from $6$ distinct Bitcoin classes: Exchange, Gambling, Marketplace, Mining Pool, Mixer and Service. Here, our experiments are predominantly focused on one critical Bitcoin class - the one that presented with the smallest number of samples in the address behavioural dataset - the Mining Pool.

The rest of the paper is organized as follows. Section~\ref{sec:related} describes related work. Section~\ref{sec:methodology} introduces GAN concepts and how they were implemented in this paper. Section~\ref{sec:experimentFramework} shows an overview of the used datasets and presents our experiments. Section~\ref{sec:results} describes the obtained results and finally, in Section~\ref{sec:conclusions}, we draw conclusions and provide guidelines for future work.

%%%%%%%%%%%%%%%%%%%%%%%%%%%%%%%%%%%%%%%%%%%%%%%%%%%%%%%%%%%%%%%%%%%%%%%%%%%%%%%%

\section{Related work}\label{sec:related}

%%%%%%%%%%%%%%%%%%%%%%%%%%%%%%%%%%%%%%%%%%%%%%%%%%%%%%%%%%%%%%%%%%%%%%%%%%%%%%%%
\subsection{Bitcoin entity recognition and class imbalance problem}\label{sec:btcentity}

To date, analysis of the Bitcoin network is predominantly focused on entity classification with the aim of detecting illicit or abnormal activities and relating them with cyber-security threats. However, often there is only little information available regarding certain actors of the Bitcoin network and critical classes are underrepresented making robust classification of all known entities difficult. Generally, imbalanced class distributions are known to hinder classifier learning \cite{fernandez2018learning}. 
%This phenomenon occurs when one or more categories (or classes) in the dataset have more samples than the others, making it hard to discover robust patterns for under-represented classes when applying supervised machine learning techniques.

Harlev et al. \cite{harlev2018breaking} applies a supervised machine learning algorithm to predict the type of yet-unidentified entities, resolving the imbalance problem by using Synthetic Minority Over-Sampling Technique (SMOTE) to over-sample underrepresented classes. However, results show that the model struggles with classes that have a low number of samples.
In \cite{ranshous2017exchange}, a single entity (exchange) is detected by extracting features related to the transaction directed hypergraph. In this case, the authors randomly sample an equal number of labeled and unlabeled addresses for training and testing their model, repeating the process $10$ times. Zola et al. \cite{zola2019cascading} implements a cascading machine learning model to detect Bitcoin entities belonging to $6$ classes, but does not consider the unbalance of the data. In \cite{lin2019evaluation}, new features for Bitcoin address classification are introduced addressing the imbalance problem using stratified random sampling. The best results show a general increase in entity classification except for two classes (Faucet and Market).
Liang et al. \cite{liang2019targeted} present an algorithm based on network representation learning to train their address multi-classifiers with imbalanced data. Bartoletti et al. \cite{bartoletti2018data} experiment with several machine learning algorithms to detect the Bitcoin Ponzi scheme using two literature approaches to solve the class imbalance problem: a cost-sensitive \cite{zhang2013cost} and a sampling-based approach \cite{chawla2004editorial}.

Since GANs have recently shown very promising results in the image processing domain \cite{wang2017generative}, we sought to investigate their potential for solving the class imbalance problem in Bitcoin entity classification by generating additional synthetic data.

\subsection{Generative Adversarial Network (GAN) applications}\label{sec:ganapp}

Generative Adversarial Networks (GANs) are an approach of generative modelling using deep learning methods such as convolutional neural networks (CNN). In \cite{goodfellow2014generative}, the potential of a GAN estimation approach is presented through a qualitative and quantitative evaluation of the generated samples.

The majority of GAN studies use images as input data, for example, in \cite{karras2017progressive} a new training methodology for creating a GAN that can generate realistic photographs of human faces is presented. Minaee et al. \cite{minaee2018finger} presents a machine learning framework based on GANs, which is able to generate fingerprint images; while in \cite{isola2017image}, authors investigate conditional adversarial networks as a solution for the image-to-image translation problem transferring style from one image to another. 
%In \cite{zhang2018stackgan++}, researchers propose a Stacked Generative Adversarial Networks (StackGANs) to generate realistic images conditioned on text descriptions. 
An interesting application is presented in \cite{reed2016generative}, where a model able to transform detailed textual descriptions of birds and flowers into images is implemented. 

Recently, GANs have been used to perform data augmentation as well. In \cite{lei2019generative} and in \cite{douzas2018effective}, GANs, respectively the Imbalanced Fusion GAN (IGAFN) and a conditional version (cGAN), are designed to approximate the true data distribution and to generate data for the minority class of various imbalanced datasets. In \cite{mariani2018bagan}, the presented Balancing GAN (BAGAN) methodology aims to generate realistic minority-class images, for example.
%Medical imaging datasets are often imbalanced as pathological findings are generally rare. 
In \cite{shin2018medical}, a method to generate synthetic abnormal magnetic resonance images (MRI) with brain tumors using a GAN approach is presented. 
%The method is based on training a GAN using two public datasets of MRI data. 
Similarly, Frid et al. \cite{frid2018synthetic} apply GANs to show that a classifier trained with synthetic images of liver lesions achieves better values of sensitivity and specificity than a classifier implemented with an imbalanced dataset. Other examples include breast cancer classification as shown in \cite{wu2018conditional}, where a conditional GAN is implemented in order to synthesize lesions from real mammogram images.

%A framework called CycleGAN for increasing the number of labeled examples in an emotion database (facial expression images) is presented in \cite{zhu2017data}. These synthetic examples added to the initial dataset generate an increase in the final classification accuracy.
In the field of cyber-security, Merino et al. \cite{merino2019expansion} suggest that GANs are a viable approach for improving cyber-attack intrusion detection systems. They generate new cyber-attack data from existing data with the goal of balancing the datasets.

Based on the promising results in the image processing domain and - recently - in other domains, we propose here to investigate how GANs can be applied to generate synthetic behavioural data of a specific, typically underrepresented Bitcoin entity. In particular, several GAN configurations are tested in order to determine the parameter setting that should be used to generate "good" quality synthetic samples with high similarity between synthetic and real data.

To the best of our knowledge, this is the first work that explores the use of GANs for Bitcoin data augmentation.

%%%%%%%%%%%%%%%%%%%%%%%%%%%%%%%%%%%%%%%%%%%%%%%%%%%%%%%%%%%%%%%%%%%%%%  

\section{Methodology}\label{sec:methodology}

\subsection{GAN overview}\label{sec:gan}
Generative modelling is an unsupervised learning task in the field of machine learning introduced by \cite{goodfellow2014generative}. A GAN is composed of two networks that compete with each other, thereby increasing their ability to learn from each other. The aim of GANs is to discover and learn regularities and patterns present in input data and generate new synthetic samples with high similarity to the original/real dataset.

A GAN is composed by two concurrent neural networks: a Generator (G) and a Discriminator (D). The task of the first network G is to generate synthetic samples while the second one D evaluates their authenticity. In particular, a generative model G captures the input data distribution, and a discriminative model D estimates the probability that a sample came from the training data rather than from G. This procedure is repeated for a certain number of times, named epochs. During each epoch, cost functions are calculated, and according to these values the weights of the two networks are changed. The idea is to minimize discriminator mistakes increasing the similarity between the synthetic (fake) and real (original) samples.

As introduced by Goodfellow et al. \cite{goodfellow2014generative}, GAN training is characterized by a game between two loss functions: one that involves the discriminator D and the other that involves the generator G. This problem is typically represented as a min-max optimization problem (Equation \ref{eq:optimizer}).

\begin{equation}\label{eq:optimizer}
\begin{split}
\min_{G}\max_{D} V(D,G) & =  E_{x \sim p_{data}(x)}[log D(x)] \\& +E_{z \sim p_{z}(z)}[log(1-D(G(z))]
\end{split}
\end{equation}

During the training phase, using gradient descent optimization, both G and D are updated simultaneously through stochastic gradient updates. Yet, this method could be affected by several issues that can cause non-convergence of the solution. Local minima and saddle points, for example, can stall the training and pathological curvature can slow down training without finding the solution. 
This problem has already been addressed by many authors \cite{ruder2016overview,nagarajan2017gradient,kingma2014adam}, where each one proposes and analyzes several optimization functions. These studies conclude that each scenario has its own suitable optimization function, and that there is no ``one-fits-all" solution that works in every situation.

For the implementation of GANs in this paper, we thus aimed to find a suitable GAN configuration and compare two standard optimization methods called Root Mean Square Propagation (RMSProp) and Adaptative Moment Optimization (ADAM). Root Mean Square Propagation tries to diminish oscillations that can slow down the optimization \cite{tieleman2012lecture} and automatically adjusts the learning rate - it is able to choose a different learning rate for each parameter. In RMSProp, updates are done according to Equation \ref{eq:rmsprop} \cite{riedmiller1993direct}.

\begin{equation}\label{eq:rmsprop}
\begin{split}
& \nu_{t} = \alpha \nu_{t-1}+(1-\alpha) * g_{t}^2 \\
& \Delta \omega_{t} = - \frac{\eta}{\sqrt{\nu_{t}}+\epsilon}*g_{t} \\
& \omega_{t+1} = \omega_{t}+\Delta \omega_{t} \\
& \eta : \textrm{Initial Learning rate} \\
& \nu_{t} : \textrm{Exp. Avg. of squares of gradients}\\
& g_{t} : \textrm{Gradient at time t along }\omega^j \\
& \alpha : \textrm{Hyperparameter}
\end{split}
\end{equation}

In Equation \ref{eq:rmsprop}, $\nu_{t}$ represents the exponential average of the square of the gradient. The exponential average is useful as it helps weigh more recent gradient updates more than the less recent ones. Then, step size $\Delta \omega_{t}$ is calculated, moving in the direction of the gradient. This step size is affected by the exponential average and two parameters $\eta$ (Initial Learning rate) and $\epsilon$. Finally, the step ($\omega_{t+1}$) is updated.
The $\alpha$ hyperparameter is typically chosen to be $0.9$ and $\epsilon$ is chosen to be $1e^{-10}$.

ADAM was introduced for the first time in \cite{kingma2014adam} and, similar to RMSProp, has the aim to diminish oscillations during the gradient descent process. However, ADAM also accelerates the optimization in the direction of the minimum.

Equation \ref{eq:rmsprop} and Equation \ref{eq:adam}, show the similarity between RMSProp and ADAM. However, for ADAM (Equation \ref{eq:adam}), an additional equation is considered and the step size computation has a small variation. The additional equation is the exponential average of gradients. This equation tries to avoid zig-zag directions. ADAM computes step size by multiplying the exponential average of the gradient as well.
Regarding the hyperparameters, $\beta_{1}$ is typically chosen to be $0.9$, $\beta_{2}$ is kept around $0.999$ and $\epsilon$ is chosen to be $1e^{-10}$.

For training the GAN, it is important fix the size of the initial dataset and choose the size of the batch. This batch size, in fact, represents the number of elements of the real dataset and the synthetic dataset used at once to update the weight of the generator and discriminator network \cite{goodfellow2016nips}.

\subsection{GAN implementation}\label{sec:ganimplementation}

The aim of this study is to determine the adequate configuration of a GAN able to recreate synthetic Bitcoin address behaviour. The proposed approach may be used to resolve Bitcoin class imbalance problems for improved entity classification.

As described in Section \ref{sec:gan}, the GAN is composed of two networks. For the purpose of this study, a neural network with three hidden layers was used as generator G. In particular, each layer was composed by respectively $512$, $256$ and $128$ neurons, all using the Rectified Linear Unit (ReLu) activation function. 
For the discriminator D, a neural network with three hidden layers was implemented. Each layer was created by respectively $256$, $512$ and $256$ neurons.

\begin{equation}\label{eq:adam}
\begin{split}
& \nu_{t} = \beta_{1} * \nu_{t-1}+(1-\beta_{1})*g_{t} \\
& s_{t} = \beta_{2} * s_{t-1}+(1-\beta_{2})*g_{t}^2 \\
& \Delta \omega_{t} = - \eta \frac{\nu_{t}}{\sqrt{s_{t}}+\epsilon}*g_{t} \\
& \omega_{t+1} = \omega_{t}+\Delta \omega_{t} \\
& \eta : \textrm{Initial Learning rate} \\
& \nu_{t} : \textrm{Exp. Avg. of gradients along } \omega_{j}\\
& s_{t} : \textrm{Exp. Avg. of squares of gradients along} \omega_{j}\\
& g_{t} : \textrm{Gradient at time t along }\omega^j \\
&\beta_{1},\beta_{2} : \textrm{Hyperparameters}
\end{split}
\end{equation}

The objective of the two networks during the training phase is to optimize and minimize the generation loss and the discrimination loss as explained in Section \ref{sec:gan}. The tested optimization functions were RMSProp and ADAM. Both of the functions were configured with equal learning rate ($0.001$).

In order to improve the GAN learning skills, each feature belonging to the original input dataset was normalized. This normalization allowed us to limit all feature distributions in the range of $0$ to $1$. 
For each feature, the maximum and minimum were computed and Equation \ref{eq:norm} was applied.

\begin{equation}\label{eq:norm}
X_{norm} = \frac{x-X_{min}}{X_{max}-X_{min}}
\end{equation}

When the training phase starts, in each epoch, a set of $N$ samples coming from a random normal distribution were used as input to G in order to create a first version of synthetic samples. In our implementation, we decided to keep the $N$ input values fixed and equal to $100$.
Once the synthetic samples were generated, D was trained in order to distinguish the synthetic samples and the real samples. D was trained with a labelled dataset composed of $2$x$M$ elements: $M$ synthetic data with the $0-label$ and $M$ real data with the $1-label$. 
During training, each $T$ epochs, the process was stopped in order to evaluate the respective solution. Only if the solution satisfied the conditions indicated in Algorithm \ref{alg:traingan}, the training was stopped. For our purposes, $T$ was chosen equal to $1,000$ epochs.

\begin{algorithm}
\SetAlgoLined
 $epoch = 0$\;
  $training$ $GAN$\;
 $epoch = epoch + 1$\;
 \If{$epoch$ \% $T == 0$}{
    $test$ = $0$ \;
    $accuracy$ = $0$ \;
    \For{$i$=$0$ to $n$}{
        $Z$ = $G(N,M)$ \;
        $C$ = $D(Z)$ \;
        $accuracy$ = count($C$ $>$ $thr1$) / len($C$) \;
        \If{$accuracy$ $>$ $thr2$}{
         $test$ = $test$ + $1$\;
        }
    }
    \eIf{$test$ == $n$}{
         \textbf{exit training}\;
         }
    {
    \textbf{resume training}\;
    }

 }
 \DontPrintSemicolon
 \hrulefill\;
 $G$ = Generator network\;
 $D$ = Discriminator network\;
 $N$ = $100$ (size of input Gaussian samples) \;
 $M$ = batch size\;
 $T$ = $1,000$  (epochs for test) \;
 $n$ = $5$ (number of the test repetition) \;
 $thr1$ = $0.90$ (threshold for synth. classification) \;
 $thr2$ = $0.90$ (threshold for similarity real/synth.) \;
 \caption{GAN evaluation phase}
 \label{alg:traingan}
\end{algorithm}

As indicated in Algorithm \ref{alg:traingan}, during evaluation, G creates a set of $M$ synthetic samples $G(N,M)$ that subsequently was used as input for the discriminator $D(Z)$. Once D finished the classification, the number of synthetic samples detected as real samples with an accuracy greater than a threshold value ($thr1$) was normalized. This operation is reflected in Algorithm \ref{alg:traingan} by count($C$ $>$ $thr1$) / len($C$), where $thr1$ was fixed to $0.90$. This test was repeated $n$ ($5$) times and the training process was stopped if in each of the $5$ tests the number of samples with accuracy higher than $thr1$ was more than $90$\% of the population ($thr2$ = $0.90$). If this condition was not verified, the GAN resumed training until the next test (in $1,000$ epochs).

\section{Experimental Framework}\label{sec:experimentFramework}

%%%%%%%%%%%%%%%%%%%%%%%%%%%%%%%%%%%%%%%%%%%%%%%%%%%%%%%%%%%%%%%%%%%%%%%%%%%%%%%%
\subsection{Blockchain data}\label{sec:blockchain}
%In this work, we propose to apply the GAN approach in order to create synthetic data that represent Bitcoin address behaviour. 
The first step was to extract a Bitcoin address dataset from the Bitcoin mainnet\footnote{https://bitcoin.org/en/glossary/mainnet}. The whole blockchain was downloaded, from the beginning until block number $570$,$000$, corresponding to blocks mined until April $3$rd $2019$, $09$:$20$:$08$ AM. The Bitcoin blockchain data were downloaded using the Bitcoin Core\footnote{https://bitcoin.org/en/download}.

The second dataset used in our analysis was obtained from WalletExplorer\footnote{https://www.walletexplorer.com/}. This platform represents a database where information about known entities and their related addresses are stored. This database is continuously updated and has been used as a ``ground truth" for many Bitcoin-related studies, such as \cite{boshmaf2019blocktag,zola2019bitcoin,ranshous2017exchange}.
Here, WalletExplorer data were divided into six classes:

\begin{itemize}
\item \textit{Exchange}: entities that allow their customers to trade among cryptocurrencies or to change cryptos for fiat currencies (or vice-versa);

\item \textit{Gambling}: entities that offer gambling services based on Bitcoin currency (casino, betting, roulette, etc.);

\item \textit{Mining Pool}: entities composed of a group of miners that work together sharing their resources in order to reduce the volatility of their returns;

\item \textit{Mixer}: entities that offer a service to obscure the traceability of their clients' transactions;

\item \textit{Marketplace}: entities allowing to buy any kind of goods or services using cryptocurrencies. Some of them potentially related to illicit activities \cite{foley2019sex};

\item \textit{Service}: entities that allow users to lend Bitcoins and passively earn interests, or allow them to request a loan.
\end{itemize}

\begin{table}[!htbp]
\center
\begin{tabular}{lcc}
\hline
\multicolumn{1}{c}{Class} & \multicolumn{1}{c}{\begin{tabular}[c]{@{}c@{}}\# Address\\ WalletExplorer\end{tabular}} & \multicolumn{1}{c}{\begin{tabular}[c]{@{}c@{}}\# Entity\\ WalletExplorer\end{tabular}} \\ \hline
\textit{Exchange} &9,947,450  &144  \\
\textit{Gambling} &3,050,899  &76  \\
\textit{Marketplace} &2,349,111  &20  \\
\textit{Mining Pool} &85,887 &27  \\
\textit{Mixer} &475,781 &37 \\
\textit{Service} &250,788  &23  \\\hline
\textbf{Total} & \textbf{16,159,916} & \textbf{327} \\ \hline
\end{tabular}
    \caption{Overview of used WalletExplorer data}
    \label{tab:walletexploreroverview}
\end{table}

As shown in Table \ref{tab:walletexploreroverview}, $327$ different entities and more than $16,000,000$ addresses were downloaded from WalletExplorer. The WalletExplorer data were combined with the whole blockchain in order to restrict the following analyses to properly labelled data only. This new, labelled "ground truth" dataset allowed us to create the Bitcoin address dataset and implement supervised machine learning based on known data.

Following indications provided in \cite{zola2019cascading}, features related to each distinct addresses were extracted from the Bitcoin address dataset. The created address dataframe was composed of $7$ features: the number of transactions in which a certain address was detected as receiver/sender, the amount of BTC received/sent from/to this address, the balance, uniqueness (if this address was used in one transaction only) and the number of siblings.

\begin{figure}[!htbp]
  \centering
    \includegraphics[width=\linewidth]{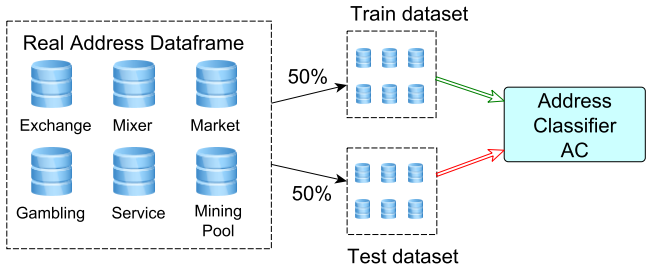}
    \caption{Address Classifier (AC) schema}
    \label{fig:address}
\end{figure}

We then split the address dataframe into two parts, as indicated in Figure \ref{fig:address} - the train and test dataset with a proportion of $50$/$50$ keeping class distributions unchanged (stratified).
The training set was used to train a machine learning model based on a Random Forest classifier creating our baseline classifier, called Address Classifier (AC). Then, the testing set was used to compute a first evaluation of how the baseline classifier trained with real data predicts the entity classes related to a certain address, as shown in Figure \ref{fig:address}.

\begin{table}[]
\center
\begin{tabular}{lccc}
\hline
\multicolumn{1}{c}{Class} & \multicolumn{1}{c}{\begin{tabular}[c]{@{}c@{}}Accuracy\\ \%\end{tabular}} & \multicolumn{1}{c}{\begin{tabular}[c]{@{}c@{}} F1-score \\ \end{tabular}} &
\multicolumn{1}{c}{\begin{tabular}[c]{@{}c@{}}Number of \\ testing samples\end{tabular}} \\ \hline
\textit{Exchange} &97.44 &0.96  &4,975,431 \\
\textit{Gambling} &87.61 &0.90  &1,523,652 \\
\textit{Marketplace} &96.46  &0.98  &1,174,643  \\
\textit{Mining Pool} &74.16 &0.81  &42,622  \\
\textit{Mixer} &79.32 &0.82  &238,056 \\
\textit{Service} &65.68  &0.73  &125,257  \\ \hline
\textbf{Total} & &  & \textbf{8,079,661} \\ \hline
\end{tabular}
    \caption{Address dataframe: obtained testing accuracy with real data}
    \label{tab:testingaccuracy}
\end{table}

Table \ref{tab:testingaccuracy} shows a summary of accuracy, f1-score and number of real samples used for testing the baseline model (AC). The created AC model yielded generally higher accuracies for detecting samples belonging to the most populated classes. In fact, the Exchange and Marketplace classes were detected with an accuracy over $96$\%, while the Service class (having overall the fewest number of samples) was detected with $65.68$\% accuracy only.
Table \ref{tab:testingaccuracy} highlights how underrepresented classes resulted in overall weak classification results.

In this study, we conducted experiments focusing only on the most underrepresented class - the Mining Pool class - and tried to generate respective synthetic samples for data augmentation. We opted to consider only the Mining Pool class as it represented the ``worst case" in terms of data due to its few number of samples (distinct addresses). The Mining Pool population was more than $100$ times smaller than the Exchange population, and almost $3$ times smaller than the Service population, and was detected with an accuracy of $74.16$\% by the baseline classifier AC.
Note that in the following sections, each time we talk about a dataset (or training dataset) we refer to a dataset composed of Mining Pool samples only.

%%%%%%%%%%%%%%%%%%%%%%%%%%%%%%%%%%%%%%%%%%%%%%%%%%%%%%%%%%%%%%%%%%%%%%%%%%%%%%%%

\subsection{GAN experiments}\label{sec:experiment}
The following experiment sought to investigate the effects of different GAN configurations. The configuration for each test was obtained by changing three relevant GAN parameters: the optimization function, the size of the input (ground-truth) dataset and the batch size used for training the network.
The two optimization functions tested in our setting were the Root Mean Square Propagation (RMSProp) and the Adaptative Moment (ADAM) as explained in Section \ref{sec:gan}.

The ground-truth dataset formed the actual input of the GAN and represented a part of the training dataset which was too big and variable to be used entirely ($ > 42,000$). In the following experiment, this dataset was generated starting from the training dataset considering Mining Pool samples only, and was normalized as described in Section \ref{sec:ganimplementation}. In particular, three ground-truth datasets were considered, each one starting from row $0$ of the Mining Pool training dataset, respectively with $10,000$, $5,000$ and $1,000$ samples. For the batch size, values of $400$, $200$, $100$ and $50$ samples were chosen.

\begin{figure}[!htbp]
  \centering
    \includegraphics[width=\linewidth]{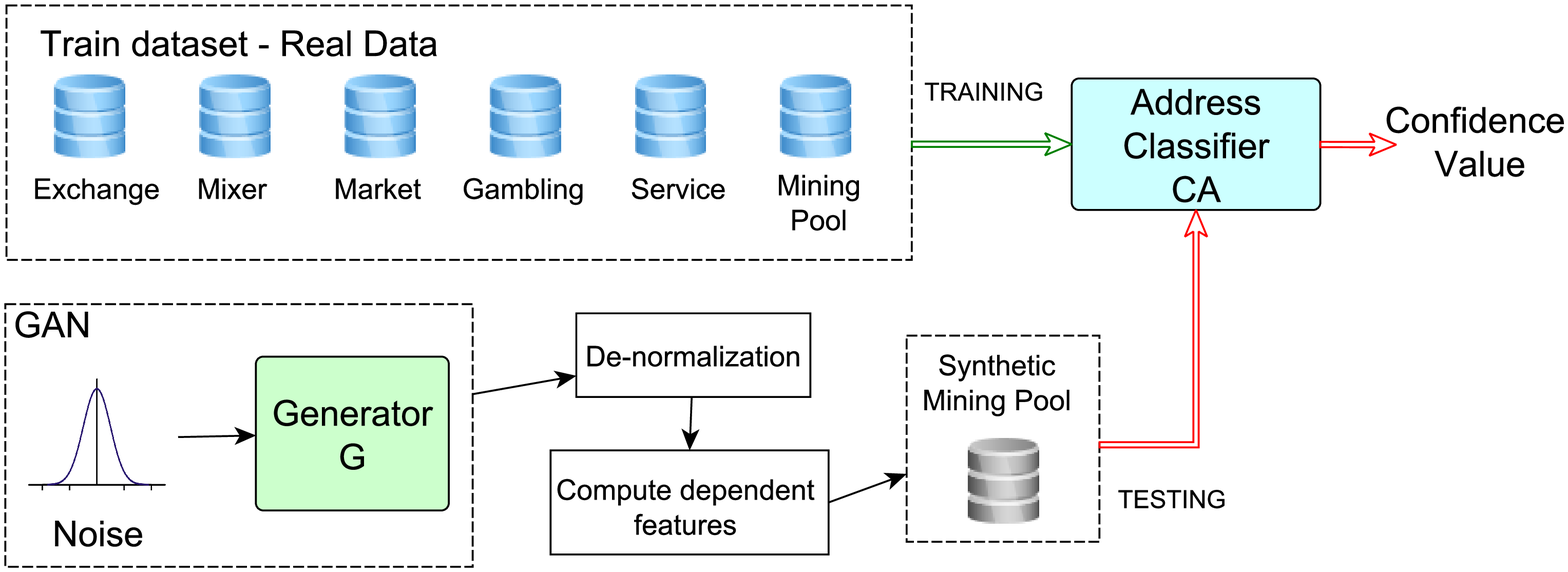}
    \caption{Experimental schema for calculating confidence values, used as measure of similarity between synthetic and real data}
    \label{fig:secondexp}
\end{figure}

Each configuration was tested three times in order to check the repeatability of the respective configuration. According to these pre-requisites, we generated $72$ different GANs, each one trained as explained in Section \ref{sec:ganimplementation}.

As the real dataset was composed of some non-independent features, the $7$ initial features were reduced to $5$, as explained in Section \ref{sec:blockchain}. In fact, the total amount of BTC received, the total amount of BTC sent and the balance are non-independent variables, so we decided to train the GAN such that it only learns two of them, while the third one was calculated. Thereby, the GAN learned the distribution of the amount of BTC received and the balance, and the amount of BTC sent was computed a-posteriori. In the same way, the uniqueness and the total received transactions are related, since one address is unique ($1$) when it is used exactly once for receiving money, otherwise is not unique ($0$). Following this rule, the total received transaction was used to train the GAN, and the uniqueness values was computed a-posteriori.

For each implementation, the number of epochs needed to train the GAN, the accuracy from D and the confidence value generated from the baseline classifier AC were calculated, as shown in Figure \ref{fig:secondexp}.
The confidence value represents the accuracy calculated via the baseline classifier, and was used here as a metric for similarity between synthetic and real samples. The accuracy from D shows the GAN's ability to cheat the discriminator D, while the confidence value evaluates the quality of the generated samples during the classification.
In order to compute the confidence value, $10,000$ synthetic samples were generated from each GAN after training. Then, these samples were de-normalized. In fact, G learned to create a distribution in the range of $0$-$1$, as explained in Section \ref{sec:ganimplementation}. This de-normalization was done by inverting Equation \ref{eq:norm} for each feature value. In this manner, the $0$-$1$ distribution was expanded to the real range of the single feature. Once the de-normalization was computed, the $2$ dependent features were computed (total amount of BTC sent and uniqueness), the obtained dataset was used to feed the AC, and to ultimately compute the confidence value (Figure \ref{fig:secondexp}).

%%%%%%%%%%%%%%%%%%%%%%%%%%%%%%%%%%%%%%%%%%%%%%%%%%%%%%%%%%%%%%%%%%%%%%%%%%%%%%%%

\section{Results}\label{sec:results}
\begin{figure}[!htbp]
  \centering
  \begin{subfigure}[b]{0.7\linewidth}
    \includegraphics[width=\linewidth]{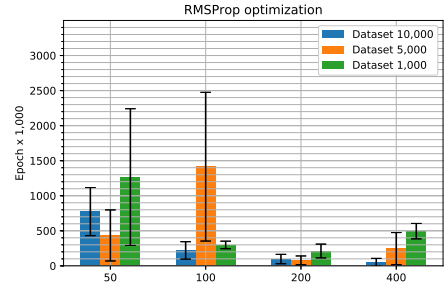}
    \caption{Epoch average for training GANs with RMSProp optimization function}
    \label{fig:epochRMSProp}
  \end{subfigure}\hspace{1em}%
  \begin{subfigure}[b]{0.7\linewidth}
    \includegraphics[width=\linewidth]{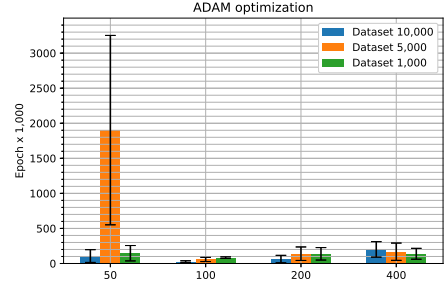}
    \caption{Epoch average for training GANs with ADAM optimization function}
    \label{fig:epochADAM}
  \end{subfigure}
  \caption{Epoch average and variance with respect to batch size in training GAN}
  \label{fig:epoch}
\end{figure}

%%%%%%%%%%%%%%%%%%%%%%%%%%%%%%%%%%%%%%%%%%%%%%%%%%%%%%%%%%%%%%%%%%%%%%%%%%%%%%%%
Figure \ref{fig:epoch} shows that the number of epochs for training a GAN using the ADAM optimizer was generally less than $200,000$ epochs. More unstable results were generated for the RMSProp optimizer, where it was not possible to establish a fixed range. Moreover, in terms of training time, the RMSProp configuration showed low repeatibility, which was reflected by its high values of standard deviation (Figure \ref{fig:epochRMSProp}). Meanwhile, the ADAM optimizer presented high variability only for the combination of batch size = $50$ and dataset size = $5,000$, which can be considered as an outlier (Figure \ref{fig:epochADAM}).

Generally, the GANs implemented with the ADAM optimizer reached a solution faster than the ones using the RMSProp optimizer, no matter what values of batch size and dataset size were chosen, as shown in Figure \ref{fig:epoch}.

\begin{table}[!htbp]
\begin{tabular}{cc|cc|cc|cc}
\multicolumn{1}{l}{} &  & \multicolumn{2}{c|}{\begin{tabular}[c]{@{}c@{}}Dataset\\ 10,000\end{tabular}} & \multicolumn{2}{c|}{\begin{tabular}[c]{@{}c@{}}Dataset\\ 5,000\end{tabular}} & \multicolumn{2}{c}{\begin{tabular}[c]{@{}c@{}}Dataset\\ 1,000\end{tabular}} \\ \hline
\# Test set & \begin{tabular}[c]{@{}c@{}}Batch\\ size\end{tabular} & \begin{tabular}[c]{@{}c@{}}Acc.\\ D\end{tabular} & \begin{tabular}[c]{@{}c@{}}Acc.\\ AC\end{tabular} & \begin{tabular}[c]{@{}c@{}}Acc.\\ D\end{tabular} & \begin{tabular}[c]{@{}c@{}}Acc.\\ AC\end{tabular} & \begin{tabular}[c]{@{}c@{}}Acc.\\ D\end{tabular} & \begin{tabular}[c]{@{}c@{}}Acc.\\ AC\end{tabular} \\ \hline

1 & 400 &1.0  &0.48  &0.99  &0.03  &1.0  &0.95  \\
2 & 400 &1.0  &0.23  &1.0  &0.08  &0.99 &0.44  \\
3 & 400 &0.92  &0.06  &0.96  &0.13  &0.97  &0.63  \\ \hline
1 & 200 &1.0  &0.07  &1.0  &0.23  &1.0  &0.18  \\
2 & 200 &1.0  &0.07  &0.99  &0.56  &1.0  &0.12  \\
3 & 200 &1.0  &0.09  &1.0  &0.04  &1.0  &0.26 \\ \hline
1 & 100 &1.0  &0.06  &1.0  &0.07  &1.0  &0.08  \\
2 & 100 &1.0  &0.09  &0.99  &0.04  &1.0  &0.13  \\
3 & 100 &0.98  &0.08  &1.0  &0.73  &1.0  &0.13  \\ \hline
1 & 50 &0.97  &0.13  &0.99  &0.01  &1.0  &0.04  \\
2 & 50 &1.0  &0.86  &1.0  &0.01  &1.0  &0.06  \\
3 & 50 &1.0  &0.12  &0.95  &0.14  &0.92  &0.05  \\ \hline
\end{tabular}
 \caption{Comparison between the accuracy computed by the discriminator D and the respective accuracy from the AC classifier fed with 10,000 samples generated from GANs with RMSProp optimizer} \label{tab:train_rms}
\end{table}

\begin{table}[!htbp]
\begin{tabular}{cc|cc|cc|cc}
\multicolumn{1}{l}{} &  & \multicolumn{2}{c|}{\begin{tabular}[c]{@{}c@{}}Dataset\\ 10,000\end{tabular}} & \multicolumn{2}{c|}{\begin{tabular}[c]{@{}c@{}}Dataset\\ 5,000\end{tabular}} & \multicolumn{2}{c}{\begin{tabular}[c]{@{}c@{}}Dataset\\ 1,000\end{tabular}} \\ \hline
\# Test set & \begin{tabular}[c]{@{}c@{}}Batch\\ size\end{tabular} & \begin{tabular}[c]{@{}c@{}}Acc.\\ D\end{tabular} & \begin{tabular}[c]{@{}c@{}}Acc.\\ AC\end{tabular} & \begin{tabular}[c]{@{}c@{}}Acc.\\ D\end{tabular}& \begin{tabular}[c]{@{}c@{}}Acc.\\ AC\end{tabular} &  \begin{tabular}[c]{@{}c@{}}Acc.\\ D\end{tabular} & \begin{tabular}[c]{@{}c@{}}Acc.\\ AC\end{tabular} \\ \hline

1 & 400 &0.99  &0.05  &0.91  &0.57  &1.0  &0.77  \\
2 & 400 &0.97  &0.03  &0.99  &0.18  &0.98 &0.98  \\
3 & 400 &1.0  &0.40  &0.95  &0.86  &0.98  &0.60  \\ \hline
1 & 200 &0.95  &0.34  &0.94  &0.13  &0.99  &0.74  \\
2 & 200 &0.99  &0.32  &0.94  &0.79  &1.0  &0.54 \\
3 & 200 &1.0  &0.21  &1.0  &0.63  &0.99  &0.66  \\ \hline
1 & 100 &1.0  &0.13  &0.93  &0.36  &0.95  &0.67  \\
2 & 100 &1.0  &0.52  &0.91  &0.13  &1.0  &0.57  \\
3 & 100 &0.93  &0.06  &0.95  &0.32  &0.98  &0.87  \\ \hline
1 & 50 &0.93  &0.20 &1.0  &0.03  &0.91  &0.09  \\
2 & 50 &1.0  &0.06  &1.0  &0.01  &1.0  &0.06 \\
3 & 50 &0.98  &0.16 &0.98  &0.06  &1.0  &0.12  \\ \hline
\end{tabular}
 \caption{Comparison between the accuracy computed from the discriminator D and the respective accuracy from the AC classifier fed with 10,000 samples generated from GANs with ADAM optimizer}
 \label{tab:train_adam}
\end{table}

Table \ref{tab:train_rms} shows the discriminator D accuracy and the baseline AC accuracy computed based on an input of $10,000$ samples generated by each configuration of GANs that use RMSProp, whereas Table \ref{tab:train_adam} shows classification results of $10,000$ samples generated by GANs implemented using the ADAM optimizer.

\begin{figure*}[!htbp]
  \centering
  \begin{subfigure}[b]{0.15\linewidth}
    \includegraphics[width=\linewidth]{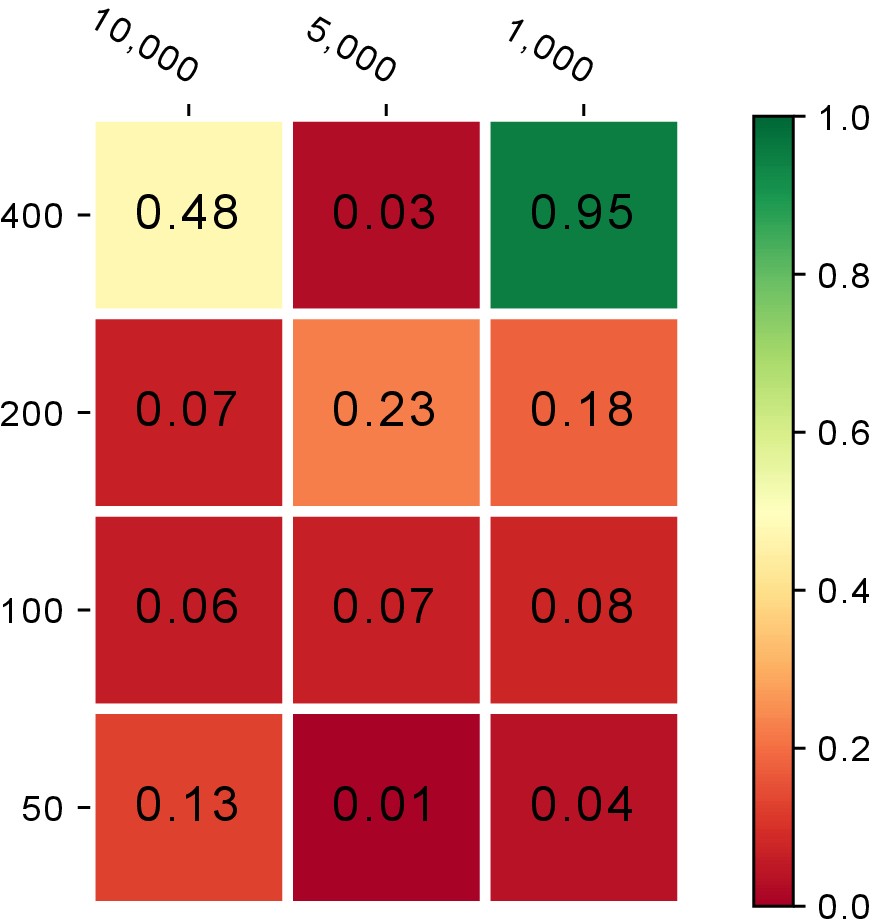}
    \caption{RMSProp first repetition}
    \label{fig:RMSProp1}
  \end{subfigure}\hspace{1em}%
  \begin{subfigure}[b]{0.15\linewidth}
    \includegraphics[width=\linewidth]{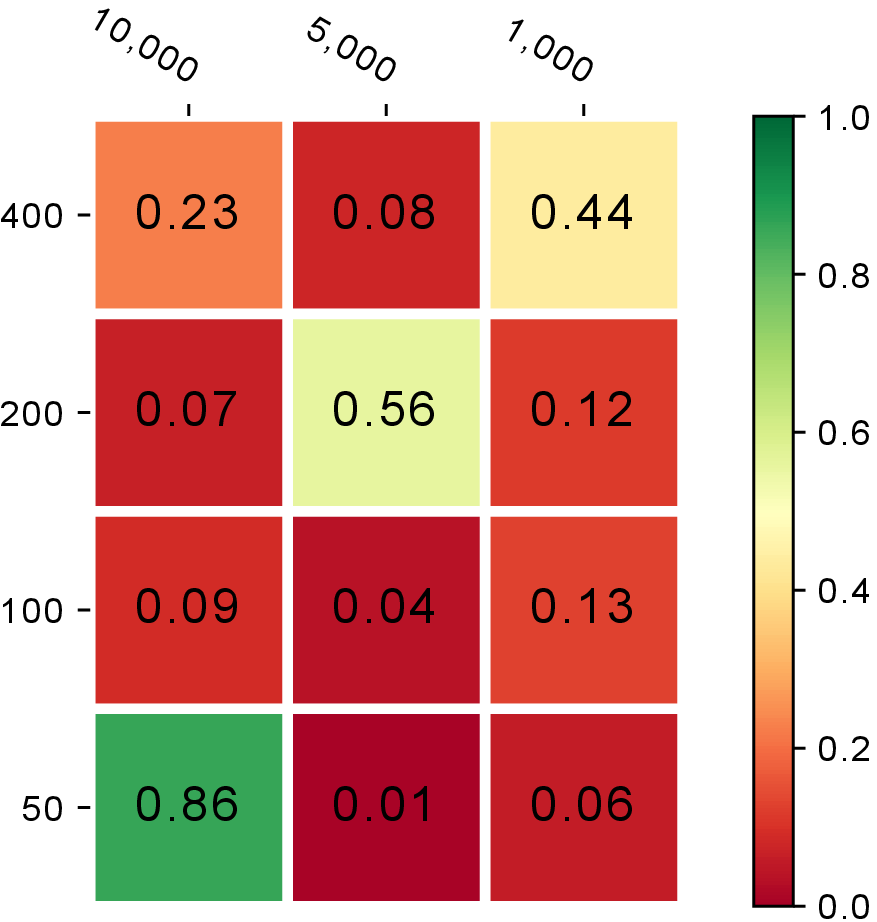}
    \caption{RMSProp second repetition}
    \label{fig:RMSProp2}
  \end{subfigure}\hspace{1em}%
  \begin{subfigure}[b]{0.15\linewidth}
    \includegraphics[width=\linewidth]{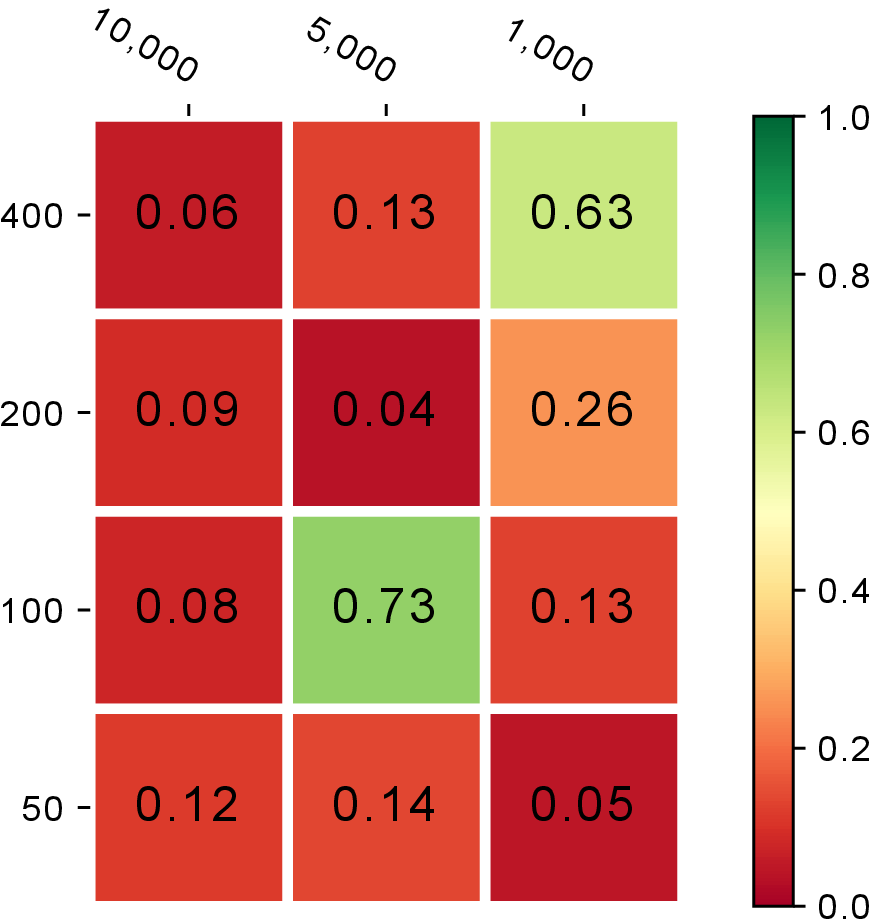}
    \caption{RMSProp third repetition}
    \label{fig:RMSProp3}
  \end{subfigure}
  \medskip
  \begin{subfigure}[b]{0.15\linewidth}
    \includegraphics[width=\linewidth]{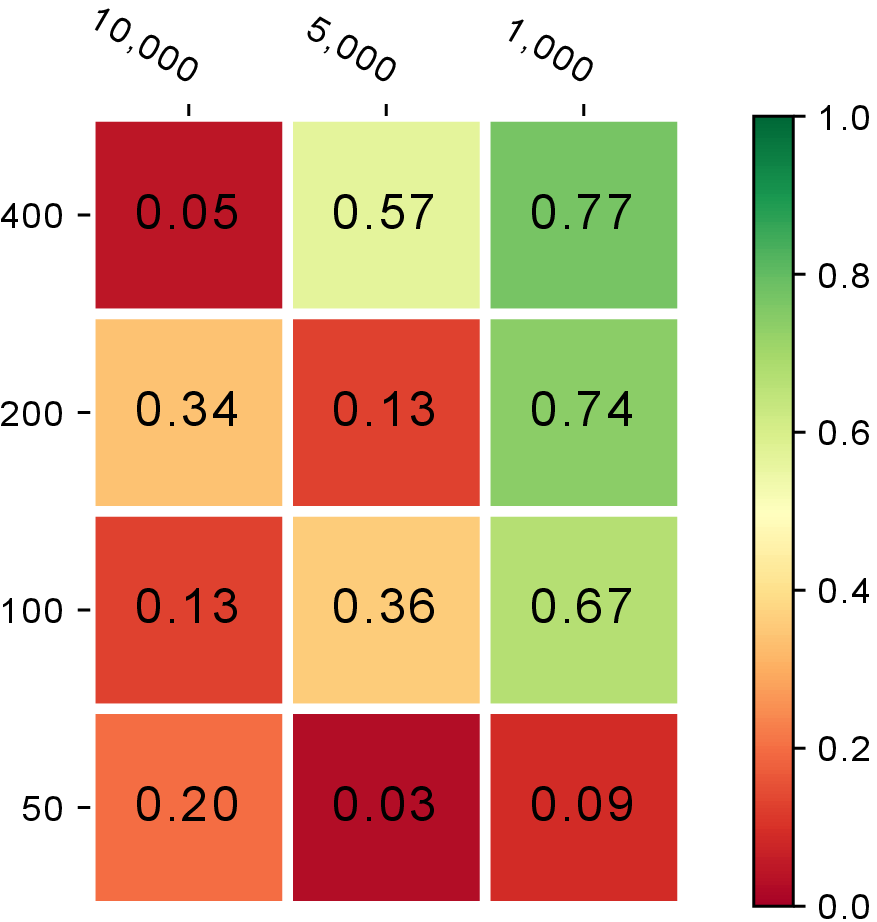}
    \caption{ADAM first repetition}
    \label{fig:ADAM1}
  \end{subfigure}\hspace{1em}%
  \begin{subfigure}[b]{0.15\linewidth}
    \includegraphics[width=\linewidth]{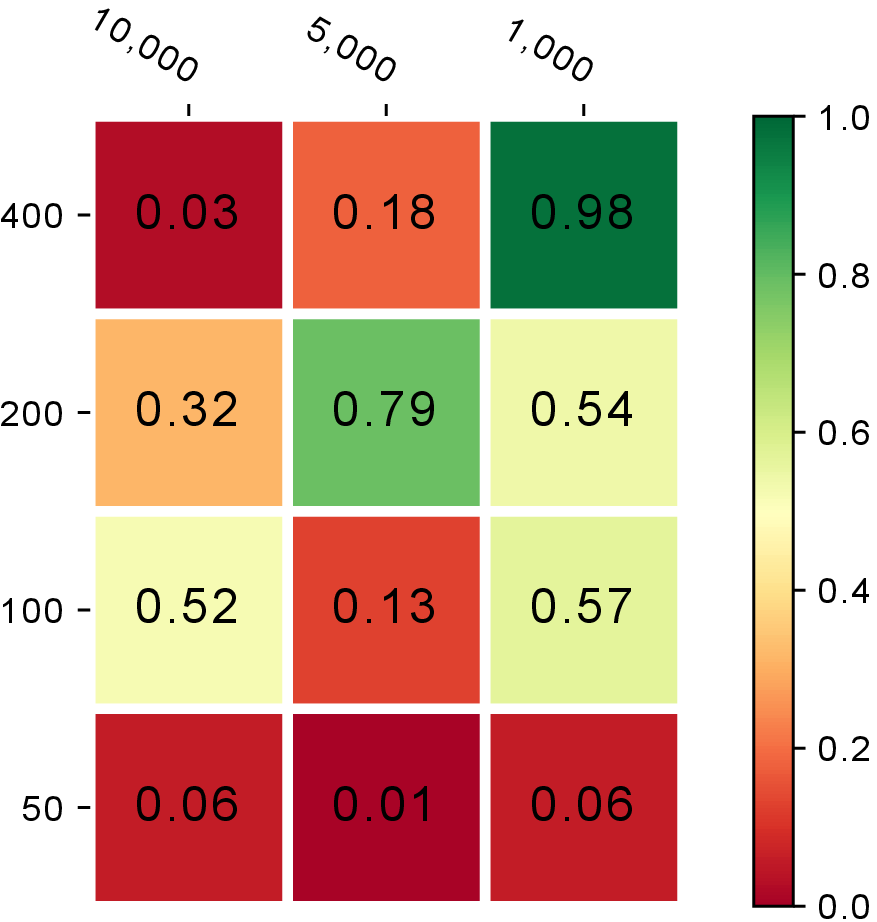}
    \caption{ADAM second repetition}
    \label{fig:ADAM2}
  \end{subfigure}\hspace{1em}%
  \begin{subfigure}[b]{0.15\linewidth}
    \includegraphics[width=\linewidth]{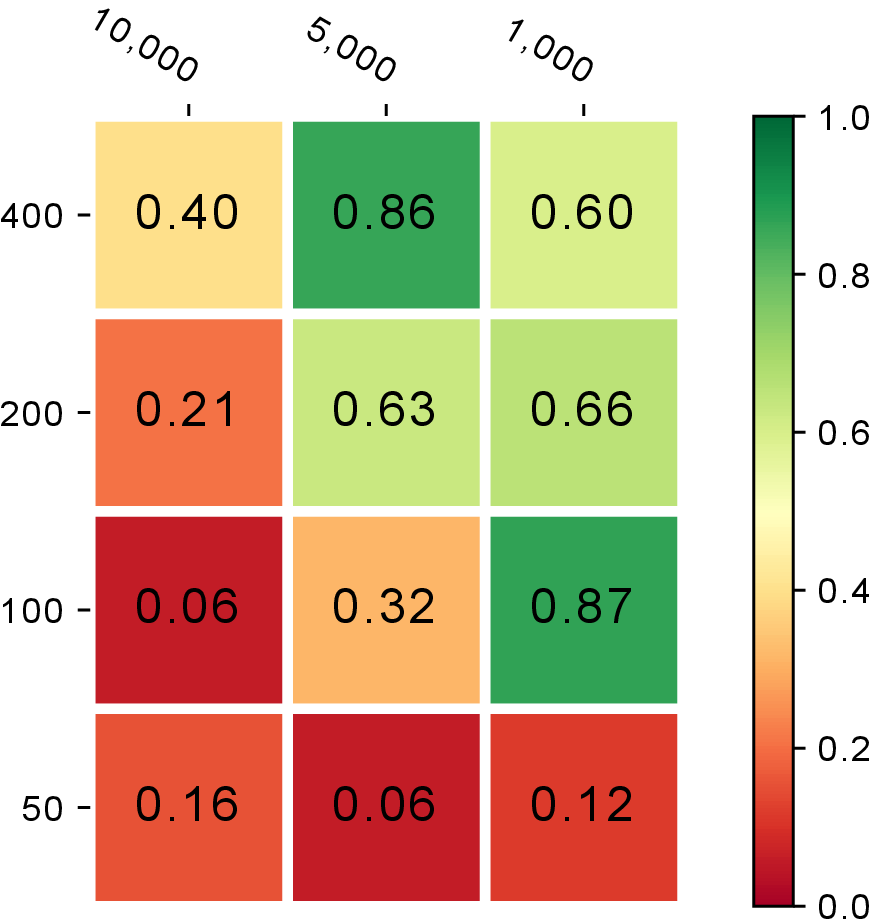}
    \caption{ADAM third repetition}
    \label{fig:ADAM3}
  \end{subfigure}
  \caption{Confidence values (measure for similarity between synthetic and real data) calculated with the baseline model AC fed with $10,000$ synthetic samples from GANs. X-axis shows influence of dataset size; Y-axis shows influence of batch size.}
  \label{fig:confidencevalues}
\end{figure*}

It is evident from Table \ref{tab:train_rms} that although RMSprop implementations presented high values of discriminator D accuracy, they showed generally poor confidence values computed with the baseline classifier (AC) based on real data. This situation is highlighted in Figure \ref{fig:RMSProp1}-\ref{fig:RMSProp3}, which shows that the similarity between the generated synthetic data and the real data used for training AC is very low. In fact, Figure \ref{fig:RMSProp1}-\ref{fig:RMSProp3} shows only a few high values but with low repeatibility. The best configuration for the RMSProp optimizer was obtained with dataset size = $1,000$ and batch size = $400$, reaching confidence values of $0.95$, $0.44$ and $0.63$ (respectively with discriminator D accuracy of $1.0$, $0.99$ and $0.97$).

Figure \ref{fig:ADAM1}-\ref{fig:ADAM3} highlights the benefit of using the ADAM classifier. In fact, in this case, more solutions presented high confidence values and good repeateability. From Figure \ref{fig:ADAM1}-\ref{fig:ADAM3} it becomes clear that the dataset size was crucial - the solutions with the best confidence values were obtained with a small dataset size ($1,000$). For the ADAM implementation, the configuration that ensured high accuracy and repeateability was obtained with the same configuration that achieved best results for RMSProp (dataset size = $1,000$ and batch size = $400$). With these settings, the GAN showed respectively $0.77$, $0.98$ and $0.60$ for AC accuracy, and $1.0$, $0.98$ and $0.98$ for discriminator D accuracy.

It is to be noted that there seemed to be a relation between the number of training epochs and the result with the best confidence values.
In Figure \ref{fig:epoch}, the best configuration with RMSProp was obtained by training the model in about $500,000$ epochs, meanwhile for the ADAM optimizer, the best configuration was achieved with only $150,000$ epochs, which is more than $3.3$ times faster.
The number of training epochs for the best ADAM configuration represented also the lowest value among the ADAM configurations with the same batch size.

%%%%%%%%%%%%%%%%%%%%%%%%%%%%%%%%%%%%%%%%%%%%%%%%%%%%%%%%%%%%%%%%%%%%%%%%%%%%%%%%
\section{Conclusions and Future work}\label{sec:conclusions}
This paper analyses how GAN configuration parameters  - optimizer function, size of the dataset and batch size - affect the training of a GAN able to generate synthetic Bitcoin address samples for data augmentation. Our approach allowed us to determine the best setting for the GAN, thus ensuring high similarity between synthetic and the real samples of a critical Bitcoin class that typically presents with few samples (Mining Pool).

Our results showed that GANs implemented with the ADAM optimizer found solutions faster than GANs using the RMSProp optimizer. Moreover, GANs using the ADAM optimizer presented high repeatibility in terms of training epochs and in terms of confidence value (baseline model accuracy testing with synthetic data; our measure for similarity between synthetic and real data), which were used as a metric for similarity between synthetic and real samples.
It is to be noted that decreasing the dataset size for GANs with the ADAM optimizer helped find solutions with high confidence values, while smaller batch size negatively affected confidence values.

Results of this study demonstrate that with the correct configuration, GANs represent a robust and efficient tool allowing for generating good synthetic data that is similar to real Bitcoin address data. In this way, GANs could be used to resolve the imbalance problem related to Bitcoin address datasets. In future work, we aim to demonstrate that the found settings generate high-quality synthetic data for other classes that are affected by the class imbalance problem (such as the Service and the Mixer) as well. Ultimately, it will be interesting to use these synthetic data to train a new general classifier and test it with real data across several classes in order to measure the effect of Bitcoin data augmentation with GANs. Based on our results, we expect a significant improvement of classification results for underrepresented classes, which could eventually improve anomaly detection within the Bitcoin network.
%On the other hand, will be interesting reconstruct the transactions related to the synthetic behaviour and simulate them in a controlled environment like the one presented in \cite{zola2019kriptosare}.

\section*{ACKNOWLEDGMENTS}
This work was partially funded by the European Commission through the Horizon 2020 research and innovation program, as part of the ``TITANIUM" project (grant agreement No 740558).

%%%%%%%%%%%%%%%%%%%%%%%%%%%%%%%%%%%%%%%%%%%%%%%%%%%%%%%%%%%%%%%%%%%%%%%%%%%%%%%%
\bibliographystyle{splncs04}
\bibliography{main}

\end{document}